\documentclass[a4paper]{article}
\usepackage{INTERSPEECH2019}

\usepackage{amsfonts, siunitx}
\usepackage{amsopn, amsmath}
\usepackage{graphicx,caption,lipsum}
\usepackage{booktabs,multirow}
\usepackage{placeins}
\usepackage{xcolor}
\usepackage{enumitem}
\usepackage{xspace,epstopdf,cite,cleveref}
\usepackage{graphicx}
\usepackage{amssymb,bm}
\usepackage{textcomp}
\usepackage{mathrsfs}  
\usepackage{xspace}

\def\vec#1{\ensuremath{\bm{{#1}}}}

\newcommand{\etal}{\textit{et al.\@}}
\newcommand{\etc}{\textit{etc.\@}}
\newcommand{\ie}{\textit{i.e.,\ }}

\newcommand{\eg}{\textit{e.g.,\ }}
\newcommand\blfootnote[1]{%
  \begingroup
  \renewcommand\thefootnote{}\footnote{#1}%
  \addtocounter{footnote}{-1}%
  \endgroup
}

\title{Modeling Interpersonal Linguistic Coordination in Conversations using Word Mover's Distance} 

\name{Md Nasir$^1$, Sandeep Nallan Chakravarthula$^1$, Brian Baucom$^2$, David C Atkins$^3$,\\ Panayiotis Georgiou$^1$, Shrikanth Narayanan$^1$}
\address{
  $^1$University of Southern California, Los Angeles, CA, USA\\
  $^2$University of Utah, Salt Lake City, UT, USA \\
  $^3$University of Washington, Seattle, WA, USA}
\email{\{mdnasir,nallanch\}@usc.edu, brian.baucom@utah.edu, datkins@uw.edu, \\
georgiou@sipi.usc.edu, shri@ee.usc.edu}

\begin{document}

\maketitle
\begin{abstract}
Linguistic coordination is a well-established phenomenon in spoken conversations and often associated with positive social behaviors and outcomes.
While there have been many attempts to measure lexical coordination or entrainment in literature, only a few have explored coordination in syntactic or semantic space.
In this work, we attempt to combine these different aspects of coordination into a single measure by leveraging distances in a neural word representation space.
In particular, we adopt the recently proposed Word Mover's Distance with \textit{word2vec} embeddings and extend it to measure the dissimilarity in language used in multiple consecutive speaker turns.
To validate our approach, we apply this measure for two case studies in the clinical psychology domain.
We find that our proposed measure is correlated with the therapist's empathy towards their patient in Motivational Interviewing and with affective behaviors in Couples Therapy.
In both case studies, our proposed metric exhibits higher correlation than previously proposed measures.
When applied to the couples with relationship improvement, we also notice a significant decrease in the proposed measure over the course of therapy, indicating higher linguistic coordination.
\end{abstract}


\section{Introduction}
When people engage in conversations in social settings, they tend to coordinate with each other and show similar behavior in various modalities.
This tendency, known as entrainment or coordination,  is exhibited through facial expressions~\cite{ramseyer2014nonverbal}, head-motion~\cite{xiao2013head}, vocal patterns (vocal entrainment)~\cite{nasir2018_towards-an-unsu, lee2014computing}, as well as the use of language (linguistic coordination)~\cite{pickering2004toward}.
Linguistic coordination is a well-established phenomenon in both spoken and written communication that has many collaborative benefits.
It is often associated with a wide range of positive social behaviors and outcomes, such as task success in collaborative games~\cite{nenkova2008high, lopes2015rule}, building effective dialogues~\cite{porzel2006entrainment} and rapport~\cite{cassell2007coordination}, engagement in tutoring scenario~\cite{ward2007_measuring-conve}, successful negotiation~\cite{taylor2008linguistic} \etc 

Understanding linguistic coordination and quantifying it is  beneficial in characterization of interpersonal behavior in psychotherapy, and in monitoring  the quality and efficacy of therapy~\cite{narayanan2013behavioral, koole2016synchrony}. 
Another potential application lies in spoken dialog systems and  conversational agents, where the system can learn to use linguistic coordination  to communicate efficiently with the human user and create a common ground~\cite{lopes2015rule}.

According to Pickering and Garrod's model~\cite{pickering2004toward}, there exist several different components in linguistic coordination -- lexical, syntactic and semantic.
Among these lexical entrainment has been arguably the focus of the most attention, primarily in  psycholinguistics~\cite{garrod1987saying, brennan1996lexical}. 
While it is a complex and multifaceted phenomenon, a number of studies have explored specific forms of lexical entrainment, such as linguistic style matching~\cite{niederhoffer2002linguistic}, similarity in choice of high frequency words~\cite{nenkova2008high}, similarity in referring expressions~\cite{brennan1996lexical}, similarity in style words~\cite{danescu2011mark} \etc \
Researchers in computational linguistics also tried to quantitatively measure lexical entrainment in conversational settings.
For example, \cite{nenkova2008high} used a unigram model of different classes of words and measured lexical entrainment as the cumulative difference in unigram scores for the interlocutors.

However, the majority of the computational approaches for measuring linguistic coordination has been limited to lexical entrainment, agnostic to coordination in the semantic space or syntactic structures.
Coordination in semantics is closely related to cohesion~\cite{halliday2014cohesion}, another mechanism in linguistics which ties together different words used in continuation of a shared context.
Approaches towards quantification of cohesion primarily have been used in tasks like text classification and discourse segmentation~\cite{manabu1994word}.
In these applications, however, cohesion is defined within a document, as opposed to the cohesion between the interlocutors in dyadic conversations which we are interested in.
There have been only a few attempts to model the latter by exploring the relation between synonymous words (\eg via WordNet) used by different speakers in the domain of intelligent tutor systems~\cite{ward2007_measuring-conve, graesser2004coh}.
However, this body of work suffers from the limitation that two words might be semantically or syntactically related even without being synonyms.
Further, using any of the lexical entrainment or cohesion measures alone does not provide a complete representation of linguistic coordination.

Addressing the aforementioned limitations and drawing inspiration from the recent success of neural word embeddings, we adopt a distance measure known as Word Mover's Distance~(WMD)~\cite{kusner2015word} and extend it to compute a distance that captures linguistic coordination.
The primary novelty in our work is in jointly integrating multiple aspects of coordination into a single measure.
In our framework, we also propose to measure the coordination locally and then normalize it globally to account for the individual tendency of coordination.
We experimentally validate our measure in two case studies in the domain of clinical psychology and psychotherapy.
We explore the proposed measure in relation to the therapist's empathy towards their patient in Motivational Interviewing as well as outcome and  affective behaviors in Couples Therapy.

\section{Lexical Similarity in Conversations}
Word Mover's Distance, originally proposed as a lexical distance between two documents, is used as a building block to measure lexical distance between two interlocutors in a dyadic conversation.
In this section, first we discuss the basics of Word Mover's Distance and then propose how the distance between utterances of two interlocutors could be used as conversational distance measure that can capture lexical and semantic dissimilarity.

\subsection{Word Mover's Distance~(WMD)}
Word Mover's Distance~(WMD) was introduced by Kusner \etal~\cite{kusner2015word}, as a distance measure between text documents.
The measure is based on the concept of neural word embeddings, which provide distributed vector representations of words in a document.
Although any neural word embedding could be used in measuring WMD, it was originally proposed using one of the most popular word embeddings, \textit{word2vec}~\cite{mikolov2013distributed}.
\textit{word2vec} has been shown to contain semantic and syntactic information~\cite{mikolov2013distributed}, making WMD suitable for capturing different aspects of  linguistic coordination.
Unlike the original WMD paper, we include stop words~(which do not carry much semantic information) in our framework, in order to capture lexical entrainment patterns of using similar high-frequency and style words.
WMD is essentially a bag-of-words approach where each document is a collection of words represented as vectors in the embedding space.
In principle, it can be interpreted as the minimum transport cost to reach the embedded words in a document from the embedded words of another document.
Inherently this measure relies on the individual distances of pairs of words in the vector space, as building blocks.
For a pair of words, $w_i$ and $w_j$, the Euclidean distance between their embedding vectors is computed as the first step, $\mathbf{v_i}=e(w_i)$ and $\mathbf{v_j}=e(w_j)$,
\begin{equation}
d(w_i,w_j)=\|\mathbf{v}_{i}-\mathbf{v}_{j}\|
\end{equation}
Based on this, the distance between a pair of utterances $U_1$ and $ U_2 $ is formulated as follows:
\begin{eqnarray}
\mathrm{WMD}(U_1,U_2)= \min_{\mathbf{T} \geq 0} \sum_{i=1}^{m}\sum_{j=1}^{n} \mathbf{T}_{i j}d(w_i,w_j) \\
\text{subject to} \quad
\sum_{j=1}^{n} \mathbf{T}_{i j}=\dfrac{c_{i}^{1}}{n} \quad \forall i \in\{1, \ldots, m\}, \nonumber
\\
\text{and} \quad\sum_{i=1}^{m} \mathbf{T}_{i j}=\dfrac{c_{j}^{2}}{m} \quad \forall j \in\{1, \ldots, n\}, \nonumber
\end{eqnarray}
where $m$ and $n$ are the number of unique words in $U_1$ and $U_2$ respectively and $c_{i}^{k}$ is the frequency of $w_i$ in $U_k$.
\begin{figure}[h]
  \centering
  \vspace{-0.5cm}  
  \includegraphics[trim= 6cm 6.3cm 6cm 4cm, clip,width=\columnwidth]{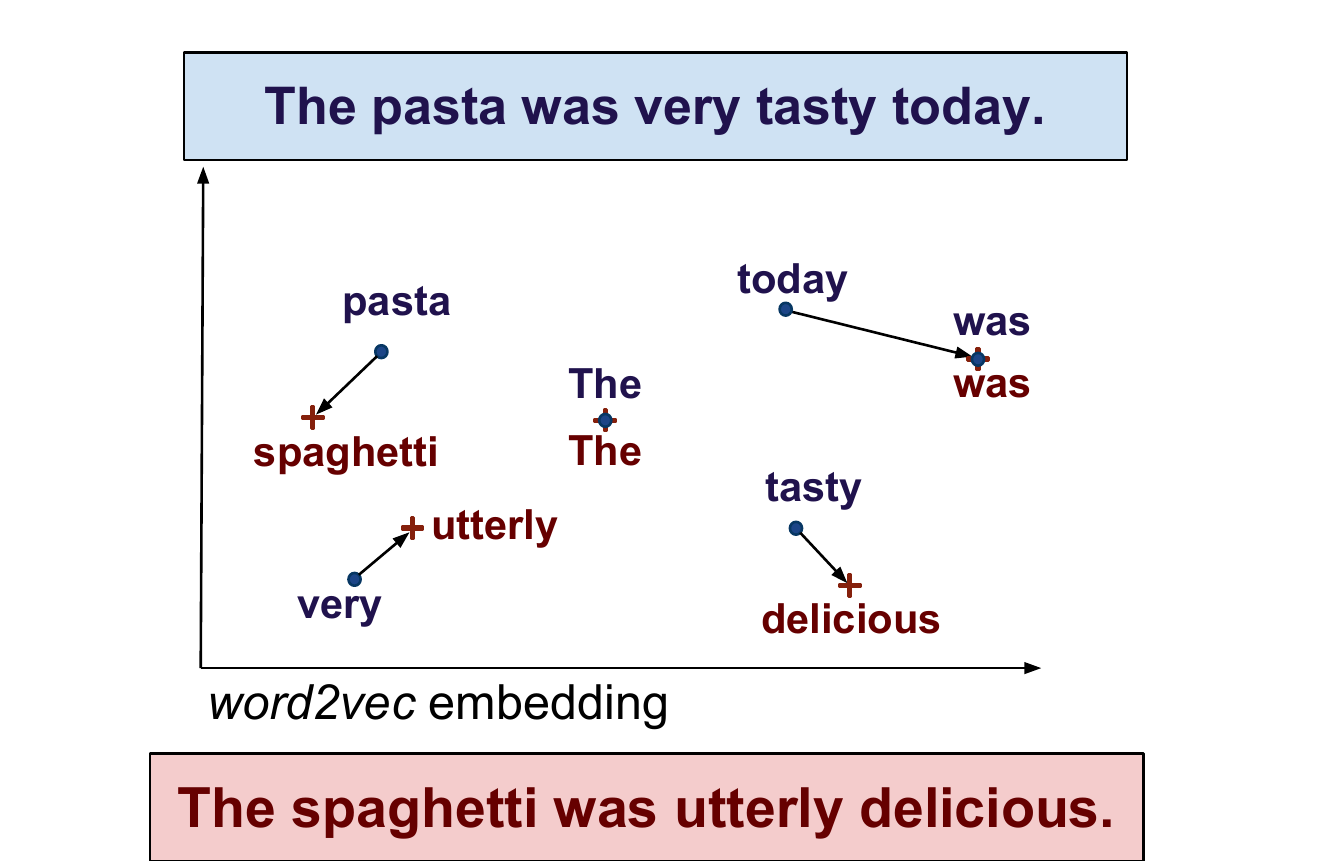}
  \caption{Illustration of WMD~(each word from one utterance is mapped to the most similar  word in the other utterance)}

\vspace{-0.4cm}
  \label{fig:wmd}
\end{figure}
The computation of WMD involves a constrained optimization problem of finding an optimal flow matrix $\mathbf{T}$ which can be solved using many exact and approximate techniques. 
In fact, this is a special case of \textit{earth mover's distance} computation, a widely-known transportation problem~\cite{rubner1998metric}.
In Figure \ref{fig:wmd}, we illustrate how WMD between two utterances is computed in the vector space of word embeddings (only two dimensions are shown for interpretability). The optimal selection of $\mathbf{T}$ could be interpreted as finding ties between neighboring words in the vector space, as seen in the figure.

Although WMD was originally introduced for documents, more recently it has been also applied for sentences~\cite{ren2018emotion}, and in this work, we use it for utterances.

\subsection{Conversational Linguistic Distances} 
As discussed earlier, WMD can provide a measure of linguistic difference between two utterances.
Here we describe how it is extended to the distance measure capturing linguistic coordination, which we name \textit{Conversational Linguistic Distance}~(CLiD). More specifically, we propose an \textit{unnormalized} and a \textit{normalized} distance~(uCLiD and nCLiD).

\subsubsection{Local Interpersonal Distance}
Although linguistic coordination occurs at multiple levels, we focus on capturing it at a local scale, \ie between consecutive turns of the interlocutors.
The other alternative is to measure the coordination globally  by considering all the words used by each of the interlocutors as a single document and computing the distance between them.
While similar approaches have been adopted in prior works on lexical entrainment~\cite{nenkova2008high}, the coarse resolution of such a measure can potentially fail to capture the dynamics of the  conversation. 

On the other hand, measuring the distance between one speaker turn and the immediate next one is a simple local measure which is appealing for our purpose.
However, local coordination is not necessarily expressed in the immediate response to the primary speaker's turn; rather it might be sustained and exhibited after a few turns~\cite{pickering2006alignment}.
Hence, we propose a scheme where we consider a predefined number of turns~(defined as context length) in response to the utterance of the primary speaker (referred to as anchor), and choose the minimum of the distances of every pair formed by the anchor utterance and a response.
This can be interpreted as the maximum coordination that is exhibited towards the primary speaker by their interlocutor in the causal vicinity of the original utterance.
In a similar approach, ~\cite{Reitter2006} considered a predefined time window (as opposed to fixed number of turns) as the context length to find instances of syntactic coordination.

Let us consider the scenario where two interlocutors $A$ and $B$ converse with each other and each of them takes $N$ number of turns.
$A_1, A_2,.., A_N$ and $B_1, B_2,.., B_N$ represent the utterances of $A$ and $B$ respectively.
Given a context length $k$, for every anchor utterance $A_i$, we compute a distance $d_i^{A \rightarrow B}$  over next $k$ number of utterances by $B$ following $A_i$ as follows:
\begin{equation}
d_i^{A \rightarrow B} = \min_{i \leq j \leq i+k-1 \leq N} \text{WMD}(A_i, B_j)
\label{eq:dist}
\end{equation} 

It should be noted that we obtain two sequences of directional distance measures for the entire session, $\{d_i^{A \rightarrow B}\}$ and $\{d_i^{B \rightarrow A}\}$, due to the asymmetric nature of Equation (\ref{eq:dist}).

\subsubsection{Session-level measures}
Although local distance measures provide a good characterization of the interpersonal coordination that happens throughout the course of conversation, an aggregated session-level measure obtained from the local distances could be more useful for session-level analysis in applications like behavioral analyses.
We simply take an average of the local distances defined in Equation (\ref{eq:dist}) over the whole session to compute the session-level measures, which we call \textit{unnormalized Conversational Linguistic Distance}~(uCLiD):
\begin{equation}
\text{uCLiD} = \frac{1}{N}\sum_{i=1}^{N} d_i^{A \rightarrow B} 
\label{eq:uCLiD}
\end{equation} 

In this equation, only uCLiD for $A \rightarrow B$ has been shown, that captures interlocutor $A$'s coordination with $B$; similarly $B \rightarrow A$ can be computed.
While the uCLiD measure provides how much overall linguistic coordination occurs between interlocutors in a conversation, it is also influenced by the nature of the conversation -- whether it is a structured conversation on pre-decided topic or an unrestricted spontaneous interaction, or something in between.
It can be also affected by the extent to which the interlocutors tend to use similar language in a conversation as a whole, as a result of coordinating to their own language.
To account for these phenomena, we use a normalized distance which attempts to provide a more suitable measure for applications where the nature of the conversation is not important. 
We draw inspiration from a similar approach by Jones \etal~\cite{jones2014finding}, where they compute a factor called \textit{Zelig Quotient} for normalization.
In our work, we first define a normalization factor $\alpha$, computed as the average pairwise WMD measure throughout the session,  including within and across interlocutors.
Next, the normalized distance measure, which we term as \textit{normalized Conversational Linguistic Distance}~(nCLiD) is computed by dividing uCLiD by $\alpha$, as follows:
\begin{align}
\text{nCLiD} &= \frac{\text{uCLiD}}{\alpha},\\
\text{where}~~
\alpha &= \frac{2}{N(N-1)}\sum_{i=1}^{N} \sum_{j=i+1}^{N}\text{WMD}(A_i, A_j)  \nonumber \\
 &+ \frac{2}{N(N-1)}\sum_{i=1}^{N} \sum_{j=i+1}^{N}\text{WMD}(B_i, B_j) \\
 &+ \frac{2}{N(N+1)}\sum_{i=1}^{N} \sum_{j=i}^{N}\text{WMD}(A_i, B_j) \nonumber
\label{nCLiD}
\end{align}
In the RHS of Equation (6), the first two terms are the average WMD within $A$ and within $B$,  which are related to the tendency to change their language throughout the conversation. 
The third term represents the overall tendency of each interlocutor to accommodate the other.

\section{Datasets}
Two datasets are used in this work: a corpus consisting of five independent clinical studies in addiction counseling (Motivational Interviewing corpus) and another corpus consisting of interactions of married couples undergoing marital therapy (Couples Therapy corpus).

\subsection{Motivational Interviewing corpus} 
This corpus consists of therapist-patient interactions in Motivational Interviewing (MI), a form of addiction counseling in psychotherapy.
In each interview, the aim of the therapist is to help the patient, who is seeking therapy for substance addiction, make behavioral changes by resolving ambivalence about their problems.
There are 145 interactions, in total, collected from the five clinical studies: ARC, ESPSB, ESB21, iCHAMP, HMCBI \cite{atkins2014scaling}. 
The interactions, which range from 20 minutes to an hour, take place between therapists and real patients struggling with alcohol, marijuana and poly-drug addiction.

Each interaction was recorded on tape and manually transcribed and annotated for speaker labels, turn timings, back-channels, disfluencies, etc.
In addition, each therapist was assigned an overall, session-level rating for the behavior code \emph{empathy} based on the Motivational Interviewing Treatment Integrity (MITI) \cite{moyers2003motivational} manual.
The rating was performed on a Likert Scale from 1 to 7, where low (high) values indicated low (high) levels of empathy exhibited by the therapist.

\subsection{Couples Therapy corpus}
The second dataset used in this work was collected as part of longitudinal study conducted by University of California, Los Angeles and University of Washington~\cite{christensen2004traditional}. 
134 seriously and chronically distressed heterosexual couples received therapy and participated in sessions where each spouse discussed with their partner one problem relevant to their relationship, without any therapist or research staff present.
There is a total of 574 such sessions, recorded at three different points of time over a span of two years while undergoing therapy~(before therapy, after 26 weeks and 2 years since the beginning of the therapy).
Along with audio-visual recordings, the corpus also includes manual transcripts with speaker labels of the conversations. 

For each session, both of the spouses are evaluated with 32 session-level behavioral codes using two separate coding schemes.
19 of the codes are based on the Social Support Interaction Rating System~(SSIRS) while 13 of them follow The Couples Interaction Rating System (CIRS).
All of these codes are rated by three to four trained annotators for each session on a scale from 1 to 9.
In this work, our focus lies on analyzing only two codes from the SSIRS system -- \textit{Global Positive Affect} and \textit{Global Negative Affect}.
Finally, the corpus also includes the therapy outcomes of the couples as a measure of their relationship quality relative to the beginning of the therapy.
Rated on two occasions (26 weeks and/or 2 years), which we refer to as \textit{post-therapy} sessions, the outcome is rated on a 4-point scale; 1~(deterioration), 2~(no change), 3~(partial recovery), and 4~(complete recovery).

\section{Experiments}
We applied the proposed measure in the two case studies using the datasets described in Section 3.
In this section, we describe the correlation analysis experiments conducted to indirectly validate our proposed measures.

\subsection{Baselines}
We use a number of baseline methods to compare with the proposed method:
\begin{itemize}
\item Turn-level lexical similarity based on TF-IDF\cite{liebman2016capturing},
\item Cohesion~(distance) measure based on WordNet~\cite{ward2007_measuring-conve},
\item Global WMD measured between the language of the interlocutors taken together, as described in Section 2.2.1
\end{itemize}

\subsection{Case Study 1: Empathy in Motivational Interviews}
Deemed an important interpersonal behavior in counseling-based psychotherapy, empathy has been shown to be positively associated with entrainment both in domain theory~\cite{preston2002empathy} and computational studies~\cite{xiao2013modeling, lord2015more}.
In this case study, we compute Spearman's $\rho$ correlation between the proposed linguistic coordination measures (uCLiD and nCLiD) and empathy ratings.
Due to the asymmetric nature of the proposed measure, we obtain each of these measures in two directions--the \textit{patient-to-therapist} and \textit{patient-to-therapist}.
Since empathy is a behavior expressed by therapist, intuitively it should not be affected how much coordination the patients exhibits.
As a verification, we found no significant correlation between the therapist-to-patient distance~(using nCLiD measure) and empathy~($\rho=0.0521, p=0.4344$).
Hence we consider only patient-to-therapist coordination distance, focusing only on the coordination exhibited by the therapist.
We empirically set the context length parameter of our measure as $k=6$ and use a 300-dimensional pre-trained model for \textit{word2vec} (trained on 3 million words from Google News).
We also report the $p$-values against the null hypothesis $H_0$ that there is no monotonic~(rank-ordered) association between empathy and the candidate measure.
We repeat the same procedure for the baselines as well.
\vspace{-1ex}
\begin{table}[h]
   \centering
      \fontsize{12}{12}\selectfont
   \renewcommand{\arraystretch}{1.5}
      \resizebox{0.65\linewidth}{!}{
       \begin{tabular}{c|c|c}
       \hline
       \multirow{2}{*}{Measure}  & \multicolumn{2}{c}{Spearman's correlation} \\
		\cline{2-3}
        & $\rho$  & $p$-$\mathrm{value}^{*}$ \\

       \hline
		uCLiD  &   $-0.2283$     &  $0.0103$            \\      
        nCLiD  &    $\mathbf{-0.2639}$    &  $0.0026$          \\
                    \hline  
        ${}^{\dag}$TF-IDF~\cite{liebman2016capturing} &    $0.1152$     &  $0.1675$           \\           
		WordNet~\cite{ward2007_measuring-conve} &  $-0.0952$     &  $0.2546$         \\  
	global WMD &  $-0.1710$     &  $0.0398$         \\  
       \hline
    \end{tabular}}
    \caption{Correlation between empathy and  various coordination measures   %
}
    \label{tab:empathy}
    \end{table}
\vspace{-5.1ex}
\blfootnote{*$p<0.05$ indicates statistically significant correlation}
\blfootnote{${ }^\dag$ similarity measure while other measures are distances}
 
From the results shown in Table \ref{tab:empathy}, we can observe that both the normalized and the unnormalized measure (uCLiD and nCLiD, respectively) exhibit stronger correlation than the baselines. 
We also notice the improvement from normalization as nCLiD turns out to be the most highly correlated measure.
The negative sign of the correlation values is justified for the proposed measures since we expect sessions with higher empathy to have higher coordination, and hence, lower distance.
We also observe $p$-values lower than $0.05$ indicating statistically significant association between empathy and the proposed measures.

\subsection{Case Study 2: Couples Therapy}
\subsubsection{Individual behavioral codes}
In the Couples Therapy domain, we first explore the possible association of linguistic coordination with \textit{positive} and \textit{negative} affect.
We adopt the same context length parameter value for our measures ($k=6$) and use the same baselines for comparison as used in the previous case study.
We consider the coordination exhibited by a subject~(husband or wife) with their spouse for the behavior ratings of the former.
For example, as far as the husband's \textit{positive} affective behavior is concerned, we only analyze how much the husband coordinated with respect to the wife during the session.
\begin{table}[h]
\vspace{-1ex}
   \centering
      \fontsize{12}{12}\selectfont
   \renewcommand{\arraystretch}{1.5}
      \resizebox{1.0\linewidth}{!}{
       \begin{tabular}{c|cc|cc}
       \hline
        \multicolumn{1}{l|}{\multirow{2}[0]{*}{Measure}}  & \multicolumn{2}{c|}{\textit{positive}} & \multicolumn{2}{c}{\textit{negative}}\\
		\cline{2-5}
       \multicolumn{1}{c|}{}  & $\rho$  & $p$-value*  & $\rho$  & $p$-value* \\

       \hline
              uCLiD  &     $-0.2903$       &  $\num{9.9e-5}$     &  $0.3142$     &  $\num{3.4e-8}$           \\
		      nCLiD  &      $\mathbf{-0.3068}$      &  $\num{1.2e-7}$     &  $\mathbf{0.3371}$ & $\num{2.1e-10}$       \\   
		      \hline    
${}^{\dag}$TF-IDF~\cite{liebman2016capturing}   &     $0.1542$       &  $0.0001$     &  $-0.2119$     & $\num{2e-4}$           \\
WordNet~\cite{ward2007_measuring-conve}  &      $-0.0847$       &  $0.0020$     &  $0.0952$     & $0.0005$       \\ 
 global WMD   &      $-0.1310$      &  $0.0001$     &  $0.1556$     & $0.0001$       \\ 
       \hline
    \end{tabular}}
    \caption{\emph{Correlation between various coordination measures and affective behaviors~(\textit{positive} and \textit{negative}})}
    \label{tab:code}
\vspace{-3ex}
\end{table}%

The results in Table \ref{tab:code} show that we obtain higher correlation values for our proposed measures than the baselines and that the normalized measure again exhibited the strongest correlation.
Judging by the sign of $\rho$, coordination distance is higher for subjects with lower \textit{positive} affect and lower for subjects with lower \textit{negative} affect, which is consistent with literature associating entrainment with behavior~\cite{lee2014computing}.

\subsubsection{Therapy outcome}
We hypothesize that the coordination distance between both spouses (measured by the average of husband-to-wife and wife-to-husband distances) decreases in the post-therapy session with respect to the pre-therapy if they had fully recovered~(outcome rating "4").
We conduct a paired Wilcoxon signed rank test against the null hypothesis $H_0$ that both pre- and post-therapy measures come from the same distribution.
We obtain $p=0.0125$ for uCLiD and $p=0.0181$ for the nCLiD measure.
This indicates a statistically significant ($p<0.05$) observation that the couples who had recovered also exhibited lower coordination distance, or in other words, higher linguistic coordination after therapy.

\section{Conclusion and Future Work}
In this work, we present a novel distance measure to quantify linguistic coordination in dyadic conversations.
Equipped with neural word embeddings, our proposed measure can potentially capture  different aspects of linguistic coordination (lexical, semantic and syntactic).
From the experiments performed in the two case studies, we establish the usefulness of the measure in capturing interpersonal behavioral information.
In future, we intend to study the effect of the context length parameter on our measure.
We could use more recent and potentially more powerful neural word embedding techniques (such as BERT, ELMo \etc) instead of \textit{word2vec} in a similar framework as presented in this paper.
Motivated by the efficacy of the neural word embeddings in relation to linguistic coordination, we would also like to explore models to jointly learn the embedding that encodes shared linguistic information between the interlocutors, similar to \cite{tseng2017approaching}.
We would also like to investigate linguistic coordination \textit{in-the-wild}  through ASR transcripts using embeddings such as \textit{conf2vec}~\cite{shivakumar2018confusion2vec}.
Another possible research direction is to investigate modeling a fused measure combining linguistic and vocal coordination.

\bibliographystyle{IEEEtran}

\bibliography{mybib,mendeley,references,lexical}

\end{document}